\documentclass{article}

\usepackage{graphicx} % more modern
\usepackage{subfigure}

\usepackage{natbib}

\usepackage{algorithm}
\usepackage{algorithmic}

\usepackage[accepted]{icml2013}

\usepackage{amsmath,amsfonts,amssymb}
\usepackage{appendix}

\newtheorem{theorem}{Theorem}
\newcommand{\beq}{\begin{equation}}
\newcommand{\eeq}{\end{equation}}

\newcommand{\beqa}{\begin{eqnarray}}
\newcommand{\eeqa}{\end{eqnarray}}

\newcommand{\beqan}{\begin{eqnarray*}}
\newcommand{\eeqan}{\end{eqnarray*}}

\newcommand{\eqdef}{:=}

\newcommand{\st}{\textstyle}

\newlength{\minipagewidth}
\renewcommand{\leq}{\leqslant}
\renewcommand{\geq}{\geqslant}
\renewcommand{\hat}{\widehat}

\newcommand{\Esp}{\mathbb{E}}
\newcommand{\cA}{\mathcal{A}}

\newcommand{\cS}{\mathcal{S}}

\newcommand{\cO}{\mathcal{O}}

\newcommand{\cR}{\mathcal{R}}

\newcommand{\Real}{\mathbb{R}}
\newcommand{\Nat}{\mathbb{N}}
\newcommand{\Hist}{\mathcal{H}}

\newcommand{\argmax}{\mathop{\mathrm{argmax}}}

\renewcommand{\v}[1]{\mathbf{#1}}
\newcommand{\pen}{\mathbf{pen}}
\newcommand{\lob}{\mathbf{lob}}

\renewcommand{\sp}{\mathbf{sp}}
\newcommand{\Diam}{D}

\newcommand{\cM}{\mathcal{M}}

\newcommand{\TRUE}{\textbf{true}}
\newcommand{\FALSE}{\textbf{false}}

\def\paradot#1{\vspace{0ex plus 0.5ex minus 0.5ex}\noindent{\bf{#1.}}}

\newcommand{\BLB}{\texttt{BLB}}
\newcommand{\AlgoName}{\texttt{OMS}}

\icmltitlerunning{Near-optimal State-Representation in RL}
\begin{document}

\twocolumn[
\icmltitle{Optimal Regret Bounds for  Selecting the State Representation in Reinforcement Learning}

% It is OKAY to include author information, even for blind
% submissions: the style file will automatically remove it for you
% unless you've provided the [accepted] option to the icml2013
% package.
\icmlauthor{Odalric-Ambrym Maillard}{odalricambrym.maillard@gmail.com}
\icmladdress{Montanuniversit\"at Leoben,
            Franz-Josef-Strasse 18, A-8700 Leoben, AUSTRIA}
\icmlauthor{Phuong Nguyen}{nmphuong@cecs.anu.edu.au}
\icmladdress{Australian National University and NICTA,
            Canberra ACT 0200, AUSTRALIA}
\icmlauthor{Ronald Ortner}{rortner@unileoben.ac.at}
\icmladdress{Montanuniversit\"at Leoben,
            Franz-Josef-Strasse 18, A-8700 Leoben, AUSTRIA}
\icmlauthor{Daniil Ryabko}{daniil@ryabko.net}
\icmladdress{INRIA Lille - Nord Europe,
            40 Avenue Halley, 59650 Villeneuve d'Ascq, FRANCE}

% You may provide any keywords that you
% find helpful for describing your paper; these are used to populate
% the "keywords" metadata in the PDF but will not be shown in the document
\icmlkeywords{boring formatting information, machine learning, ICML}

\vskip 0.3in
]

\begin{abstract}
We consider an agent interacting with an environment in a single stream of actions, observations, and rewards, with no reset.
This process is not assumed to be a Markov Decision Process (MDP).
Rather, the agent has several representations (mapping histories of past interactions to a discrete state space) 
of the environment with unknown dynamics,
only some of which result in an MDP. The goal is to minimize the average regret criterion against an agent who knows 
an MDP representation giving the highest optimal reward, and acts optimally in it.
Recent regret bounds for this setting are of order $O(T^{2/3})$ with an additive term constant yet exponential
in some characteristics of the optimal MDP.
We propose an algorithm whose regret after $T$ time steps is $O(\sqrt{T})$,  with all constants reasonably small.
This is optimal in $T$ since $O(\sqrt{T})$ is the optimal
regret in the setting of learning in a (single discrete) MDP.
\end{abstract}

%%%%%%%%
\section{Introduction}
%%%%%%%%

In Reinforcement Learning (RL), an agent has to learn a task through interactions with the environment.
The standard RL framework models the interaction of the agent and the environment
as a finite-state  Markov decision process (MDP).
Unfortunately, the real world is not (always) a finite-state MDP, and the learner often has to find
a suitable {\em state-representation model}: a function that maps
histories of actions, observations, and rewards provided by the environment into
a finite space of {\em states}, in such a way that the resulting process on the state space is Markovian, reducing the problem
to learning in a finite-state MDP. However, finding such a model is highly non-trivial. One can come up
with several representation models, many of which may lead to non-Markovian dynamics. Testing which one has the MDP property
one by one may be very costly or even impossible, as testing a statistical hypothesis
requires a workable alternative assumption on the environment.
This poses a challenging problem:
find a generic algorithm that, given several state-representation models only some of which result in an MDP, 
gets (on average) at least as much reward as an optimal policy for any of the Markovian representations.
Here we do not test the MDP property but propose to use models as long as they provide high enough rewards.

\paradot{Motivation}
One can think of specific scenarios where the setting of several state-representation models is applicable.
First, these models can be discretisations  of a continuous state space. Second, they may be discretisations of the parameter
space: this scenario has been recently considered \cite{Ortner12crl} for learning in a continuous-state MDP 
with Lipschitz continuous rewards and transition probabilities where the Lipschitz constants are unknown; 
the models are discretisations of the parameter space.
A simple example is when the process is a second-order Markov process with discrete observations:
in this case a model that maps any history to the last two observations is a Markov model;
a detailed illustration of such an example can be found, e.g., in Section 4 of \cite{MH09c}.
More generally, one can try and extract some high-level discrete  features from (continuous, high-dimensional) observations provided by the environment.
For example, the observation is a video input capturing a game board, different maps attempt to extract the (discrete) state
of the game, and we assume that at least one map is correct.
Some popular classes of models are context trees \cite{RAM96},
which are used to capture short-term memories, or probabilistic deterministic finite automata \cite{VTHCC05},
a very general class of models that can capture both short-term and long-term memories.
Since only some of the features may exhibit Markovian dynamics and/or be relevant, we want an algorithm able to exploit 
whatever is Markovian and relevant for learning.
For more details and further examples we refer to \cite{MMR11}.

 \paradot{Previous work} 
This work falls under the framework of providing performance guarantees on the average reward of a considered algorithm. 
In this setting,  the optimal regret of a learning algorithm in a finite-state MDP is $O(\sqrt{T})$. This is the  regret 
of \texttt{UCRL2} \cite{JOA10} and \texttt{Regal.D} \cite{BT09}.
 Previous work on this problem in the RL literature includes \cite{Kearns2002,Brafman03, SLWLL06}.
 Moreover, there is currently  a big interest in 
 finding practical state representations for the general RL problem where the environment's states and model are both unknown, 
e.g.\ U-trees \cite{RAM96}, MC-AIXI-CTW \cite{VNHUS11}, $\Phi$MDP \cite{MH09c}, and PSRs \cite{SJR04}. Another approach in which possible models
are known but need not be MDPs was considered in \cite{Ryabko2008}.

 For the problem considered in this paper, \cite{MMR11} recently introduced the \BLB\ algorithm that,
  given a finite set $\Phi$ of state-representation models,
  achieves regret of order $\sqrt{|\Phi|}T^{2/3}$ (where $|\Phi|$ is the number of models) in respect to the optimal policy associated with any  model that is Markovian.
 \BLB\ is based on uniform exploration of all representation models and uses the performance guarantees of \texttt{UCRL2} to control the amount of time spent on non-Markov models.
 It also makes use of some internal function in order to guess the MDP \textit{diameter} \cite{JOA10} of a Markov model,
 which leads to an additive term in the regret bound that may be exponential in the true diameter,
which means the order $T^{2/3}$ is only valid for possibly very large $T$.

 \paradot{Contribution}
We propose a new algorithm called \AlgoName\ (Optimistic Model Selection), that has regret of order $\sqrt{|\Phi|T}$,
thus establishing performance that is optimal in terms of $T$, without suffering from an unfavorable additive term in the bound and without
compromising the dependence on~$|\Phi|$.
This demonstrates that taking into consideration several possibly non-Markovian representation models does not significantly degrade the
performance of an algorithm, as compared to knowing in advance which model is the right one.
The proposed algorithm is close in spirit to the \BLB~algorithm.
However, instead of uniform exploration it uses the principle of ``optimism''
for model selection, choosing the model promising the best performance.

\paradot{Outline}
 Section~\ref{sec:setting} introduces the setting; Section~\ref{sec:algo} presents our algorithm \AlgoName;
 its performance is analysed in Section~\ref{sec:main}; proofs are in  Sections~\ref{sec:regret}, and Section~\ref{sec:disc} concludes.

%%%%%%%%
\section{Setting}\label{sec:setting}
%%%%%%%%

\paradot{Environment}
For each time step $t=1,2,\ldots$, let $\Hist_{t} := \cO\times(\cA\times\cR\times\cO)^{t-1}$ be the set of histories up to time $t$,
where $\cO$ is the set of observations, $\cA$ is a finite set of actions and $\cR = [0,1]$ is the set of possible rewards.
We consider the problem of reinforcement learning when the learner interacts sequentially with
some \textit{unknown} environment: first some initial observation $h_{1} = o_1 \in \Hist_{1}= \cO$ is provided to the learner,
then at any time step $t>0$, the learner chooses an action $a_t\in\cA$ based on the current history $h_{t} \in\Hist_{t}$,  
then receives the immediate reward $r_{t}$ and the next observation~$o_{t+1}$ from the environment.
Thus, $h_{t+1}$ is the concatenation of $h_{t}$ with $(a_t,r_t,o_{t+1})$.

%%%
\paradot{State representation models}
Let $\Phi$ be a set of state-representation models. A \textit{state-representation} model $\phi \in \Phi$ is a function from the set of histories
$\Hist := \bigcup_{t\geq1}\Hist_t$ to a finite set of states $\cS_{\phi}$. For a model $\phi$, the state at step $t$ under $\phi$
is denoted by $s_{t,\phi}:=\phi(h_{t})$ or simply $s_t$ when $\phi$ is clear from context. For the sake of simplicity, 
we assume that $\cS_\phi\cap\cS_{\phi'}=\emptyset$ for $\phi\neq\phi'$.
Further, we set $\cS:=\bigcup_{\phi\in\Phi} \cS_\phi$.

A particular role will be played by state-representation models that induce a \textit{Markov decision process (MDP)}.
An MDP is defined as a decision process in which at any discrete
time $t$, given action $a_t$, the probability of immediate reward $r_t$ and next
observation $o_{t+1}$, given the past
history $h_t$, only depends on the
current observation $o_t$. That is, $P(o_{t+1},r_{t}|h_ta_t) =
P(o_{t+1},r_{t}|o_t,a_t)$. Observations in this process are called
\textit{states} of the environment. We say that a state-representation model~$\phi$ is
a \textit{Markov model} of the environment, if the process $(s_{t,\phi},a_t,r_t), t\in\Nat$ is an MDP.
This MDP is denoted as $M(\phi)$. We will always assume that such MDPs are \textit{weakly communicating}, that is, for each pair of states $x_1,x_2$ there exists $k\in\Nat$ and a sequence of actions $\alpha_1,\ldots,\alpha_k\in\cA$ such that $P(s_{k+1,\phi}=x_2| s_{1,\phi}=x_1, a_1=\alpha_1,\ldots,a_k=\alpha_k)>0$. 
It should be noted that there may be infinitely many state-representation models under which an environment is Markov.

%%%%
\paradot{Problem description}
%%%%
Given a finite set $\Phi$ which includes at least one Markov model, we want to construct a strategy that performs as well as
the algorithm that knows any Markov model $\phi\in\Phi$, including its rewards and transition probabilities.
For that purpose we define for any Markov model $\phi\in\Phi$ the regret of any strategy at time~$T$, cf.~\cite{JOA10,BT09,MMR11}, as
\[
    \Delta(\phi,T) := T\rho^\star(\phi)-\sum_{t=1}^T{}r_t\,,
\]
where $r_t$ are the rewards received when following the proposed strategy and
$\rho^\star(\phi)$ is the optimal average reward in $\phi$, i.e.,
$\rho^\star(\phi) \eqdef \rho(M(\phi),\pi^\star_\phi) := \lim_{T \to \infty} \frac{1}{T}\Esp\big[\sum_{t=1}^T r_t(\pi^\star_\phi)\big]$
where $r_t(\pi^\star_\phi)$ are the rewards received when following the optimal policy $\pi^\star_\phi$ for $\phi$.
Note that for weakly communicating MDPs the optimal average reward indeed does not depend on the initial state.
One could replace $T\rho^\star(\phi)$ with the expected sum of rewards obtained in $T$ steps (following the optimal policy)
at the price of an additional $O(\sqrt{T})$ term.

%%%%
\section{Algorithm}\label{sec:algo}
%%%%

\paradot{High-level overview}
The \AlgoName~algorithm we propose (shown in detail as Algorithm~\ref{fig:algo}) proceeds in episodes $k=1,2,\ldots$, each consisting of several runs $j=1,2,\ldots$.
In each run $j$ of some episode $k$, starting at time $t=t_{k,j}$, \AlgoName~chooses a policy $\pi_{k,j}$ applying the optimism in face of uncertainty principle twice.
First, in line~6, \AlgoName~considers for each model $\phi\in\Phi$  a set of admissible MDPs $\cM_{t,\phi}$
(defined via confidence intervals for the estimates so far), and computes
a so-called optimistic MDP $M_t^+(\phi)\in \cM_{t,\phi}$ and an 
associated optimal policy $\pi_t^+(\phi)$ on $M_t^+(\phi)$ such that the average
reward $\rho(M_t^+(\phi),\pi_t^+(\phi))$ is maximized.
Then (line~7) \AlgoName~chooses the model $\phi_{k,j}\in \Phi$ which maximizes the average reward $\pi_{k,j} := \pi_t^+(\phi_{k,j})$
penalized by a term intuitively accounting for the ``complexity'' of the model,
similar to the \texttt{REGAL} algorithm of \cite{BT09}.

The policy $\pi_{k,j}$ is then executed until either (i) run~$j$ reaches the maximal length of $2^j$ steps (line~19),
(ii)~episode $k$ terminates when the number of visits in some state has been doubled (line~17), or
(iii) the executed policy $\pi_{k,j}$  does not give sufficiently high average reward (line~12).
Note that \AlgoName~assumes each model to be Markov, as long as it performs well.
Otherwise the model is eliminated (line~15).

\begin{algorithm}[!t]
\caption{Optimistic Model Selection (\AlgoName)} \label{fig:algo}
\begin{algorithmic}[1]
\REQUIRE  Set of models $\Phi_0$, parameter $\delta\in[0,1]$.
\STATE Set $t := 1$, $k:=0$, and $\Phi := \Phi_0$.
\WHILE{\TRUE}
 \STATE $k:=k+1$, $j:=1,$ sameEpisode := \TRUE
\WHILE{sameEpisode}
 \STATE $t_{k,j} := t$\\
\STATE $\forall\phi\in\Phi$, use EVI to compute optimistic MDP $M_t^+(\phi)\in \cM_{t,\phi}$ and 
(near-)optimal policy $\pi_t^+(\phi)$ with approximate optimistic average reward $\hat \rho_{t_{k,j}}^+(\phi)$.
\STATE Choose model $\phi_{k,j}\in  \Phi $ such that
\begin{equation}\label{eq:pen}
    \phi_{k,j}=\argmax_{\phi\in\Phi}\Big\{\hat \rho_{t_{k,j}}^+(\phi)- \pen(\phi;t_{k,j})\Big\}\,.
\end{equation}
\STATE Define $\rho_{k,j} := \hat \rho_{t_{k,j}}^+(\phi_{{k,j}}), \pi_{k,j} := \pi^+_{t_{k,j}}(\phi_{k,j}).$
\STATE  sameRun := \TRUE.
\WHILE{sameRun}
\STATE Choose action $a_t := \pi_{k,j}(s_t)$, get reward $r_t$, observe next state $s_{t+1} \in \cS_{{k,j}}:=\cS_{\phi_{k,j}}$.
\STATE Set testFail := \TRUE\,iff the sum of the collected rewards so far from time $t_{k,j}$ is  less than
\begin{equation}\label{eq:test}
\ell_{k,j}\rho_{k,j} - \lob_{k,j}(t),
\end{equation}
where $\ell_{k,j} \eqdef t-t_{k,j}+1$.
\IF{testFail}
 \STATE sameRun := \FALSE, sameEpisode := \FALSE
 \STATE $\Phi := \Phi \setminus \{\phi_{k,j}\}$
\STATE \textbf{if} $\Phi = \emptyset$ \textbf{then} $\Phi := \Phi_0$  \textbf{end if}
\ELSIF{$v_k(s_t,a_t)= N_{t_k}(s_t,a_t)$}
 \STATE sameRun := \FALSE, sameEpisode := \FALSE
\ELSIF{$\ell_{k,j}= 2^j$}
  \STATE sameRun := \FALSE, $j:=j+1$
\ENDIF
\STATE $t:=t+1$
\ENDWHILE
\ENDWHILE
\ENDWHILE
\end{algorithmic}
\end{algorithm}

\paradot{Details}
We continue with some details of the algorithm. In the following, $S_\phi:=|\mathcal S_\phi|$
denotes the number of states under model $\phi$, $S:=|\mathcal S|$ is the total number of states, 
and $A:=|\mathcal A|$ is the number of actions.
Further, $\delta_t:=\delta/36t^2$ is the confidence parameter for time~$t$.

\paradot{Admissible models}
First, the set of \textit{admissible} MDPs $\cM_{t,\phi}$ the algorithm considers at time~$t$ for each model $\phi\in\Phi$ is
defined to contain all MDPs with state space $\mathcal S_\phi$ and with
rewards $r$ and transition probabilities $p$ satisfying
\begin{eqnarray}
\big\|  p(\cdot | s,a)-\hat p_t(\cdot | s,a)\big\|_1 &\leq& \sqrt{\tfrac{2\log(2^{S_\phi}S_\phi A t/\delta_{t})}{N_{t}(s,a)}}\,,\quad \label{eqn:cond_p}\\
 \big| r(s,a) - \hat r_t(s,a)\big| &\leq& \sqrt{\tfrac{\log(2S_\phi A t/\delta_{t})}{2N_{t}(s,a)}}\,,\label{eqn:cond_r}
\end{eqnarray}
where $\hat p_t(\cdot | s,a)$ and $\hat r_t(s,a)$ are respectively the empirical transition probabilities and 
mean rewards (at time~$t$) for taking action $a$ at state $s$, and $N_{t}(s,a)$ is the number of times action $a$ has been chosen in state~$s$ up to time $t$.
(If $a$ hasn't been chosen in $s$ so far, we set $N_{t}(s,a)$ to 1.)
It can be shown (cf.\ Appendix C.1 of \citet{JOA10}) 
that the mean rewards~$r$ and the transition probabilities~$p$
of a Markovian state-representation $\phi$ satisfy~\eqref{eqn:cond_p} and \eqref{eqn:cond_r} at time~$t$
for all $s\in\cS_\phi$ and $a\in\cA$, each with probability at least $1-\delta_{t}$, making Markov models admissible with high probability.

\paradot{Extended Value Iteration}
For computing a near-optimal policy  $\pi_{t}^+(\phi)$ and a corresponding optimistic MDP $M_t^+(\phi)\in\cM_{t,\phi}$ (line~6),
\AlgoName~applies for each $\phi\in\Phi$ extended value iteration (EVI) \cite{JOA10} with precision parameter $t^{-1/2}$.
EVI computes optimistic approximate state values $\v{u}^+_{t,\phi} = (u^+_{t,\phi}(s))_s \in \Real^{S_\phi}$
just like ordinary value iteration \cite{puterman} with an additional optimization step for choosing the transition kernel
maximizing the average reward.
The (approximate) average reward $\hat \rho_t^+(\phi)$ of $\pi_{t}^+(\phi)$ in $M_t^+(\phi)$ 
then is given by
\begin{multline}\label{eqn:empevi}
\hat \rho_t^+(\phi) = \min\Big\{ r^+_t(s,\pi_t^+(\phi,s))\\
  +\sum_{s'} p^+_t(s'|s)\, u^+_{t,\phi}(s')-u^+_{t,\phi}(s), s\in\cS_\phi\Big\}\,,
\end{multline}
where $r^+_t$ and $p^+_t$ are the rewards and transition probabilities of $M_t^+(\phi)$
under $\pi_t^+(\phi)$.
It can be shown \cite{JOA10} that $\hat \rho_t^+(\phi) \geq \rho^\star(\phi) - 2/\sqrt{t}$.

\paradot{Penalization term}
At time $t=t_{k,j}$, we define the empirical value span of the optimistic MDP $M_t^+(\phi)$ as
$\sp(\v{u}^+_{t,\phi}):= \max_{s\in\cS_{\phi}}u^+_{t,\phi}(s)- \min_{s\in\cS_{\phi}} u^+_{t,\phi}(s)$,
and the penalization term considered in \eqref{eq:pen} for each model $\phi$ is given by
\beqan
\pen(\phi;t) &\eqdef&  2^{-j/2} \, c(\phi;t) \,\sp(\v{u}^+_{t,\phi})   \\
		&&+\, 2^{-j/2} \, c'(\phi;t)  + 2^{-j} \,\sp(\v{u}^+_{t,\phi}),
\eeqan
where the constants are given by
\beqan
c(\phi;t) \!\!&\eqdef&\!\! 2  \sqrt{2 S_{\phi} A \log(2^{S_{\phi}} S_{\phi} A t/\delta_{t})} + 2\sqrt{2\log(\tfrac{1}{\delta_{t}})}, \\
c'(\phi;t) \!\!&\eqdef&\!\! 2\sqrt{2 S_{\phi} A \log(2 S_{\phi} A t/\delta_{t})}\,.
\eeqan

\paradot{Deviation from the optimal reward}
Let $\ell_{k,j} \eqdef t-t_{k,j}+1$, and $v_{k,j}(s,a)$ be the total number of times $a$ has been played in $s$ during run $j$ in episode $k$ 
(or until current time $t$ if $j$ is the current run). Similarly, we write $v_{k}(s,a)$ for the respective total number of visits during episode $k$.
(Note that by the assumption $\cS_\phi \cap \cS_{\phi'} = \emptyset$ for $\phi \neq \phi'$, 
the state implicitly determines the respective model.)
Then for the test \eqref{eq:test} that decides whether the chosen model $\phi_{k,j}$ gives sufficiently high reward,
we define the allowed deviation from the optimal average reward in the optimistic model for any $t\geq t_{k,j}$ in run~$j$ as
\beqa
\lefteqn{\lob_{k,j}(t) \eqdef
 2\sum_{s \in \cS_{{k,j}}}\sum_{a\in\cA} \sqrt{2 v_{k,j}(s,a) \log\!\big(\tfrac{2 S_{{k,j}} A t_{k,j}}{\delta_{t_{k,j}}}\big)}}  \nonumber\\
&&  2\sp_{k,j}^{+}\!\!\!\sum_{s \in \cS_{{k,j}}}\sum_{a\in\cA} \sqrt{2 v_{k,j}(s,a)
		\log\!\big(\tfrac{2^{S_{{k,j}}} S_{{k,j}} A t_{k,j}}{\delta_{t_{k,j}}}\big)}\;  \nonumber\\
&& + \,\,2\sp_{k,j}^{+}\sqrt{2\ell_{k,j}\log(1/\delta_{t_{k,j}})} + \sp_{k,j}^{+}\,, \label{eq:lob}
\eeqa
where $\sp_{k,j}^+:=\sp(\v{u}^+_{t_{k,j},\phi_{k,j}})$ and $S_{k,j} := S_{\phi_{k,j}}$.
Intuitively, the first two terms correspond to the estimation error of the transition kernel and the rewards, 
while the last one is due to stochasticity of the sampling process.

%%%%%%%%
\section{Main result}\label{sec:main}
%%%%%%%%
We now provide the main result of this paper, an upper bound on the regret of our \AlgoName~strategy.
The bound involves the \textit{diameter} of a Markov model $\phi$, $\Diam(\phi)$, which is defined as the expected minimum time
required to reach any state starting from any other state in the MDP~$M(\phi)$ \cite{JOA10}.

\begin{theorem}\label{thm:mainFinite}
Let $\phi^\star$ be an optimal model, i.e.\  $\phi^\star\in \argmax\big\{\,\rho^\star(\phi)\,|\, \phi\in \Phi,\, \phi\text{ is Markovian}\,\big\}$.
Then the regret $\Delta(\phi^\star,T)$ of \AlgoName~(with parameter $\delta$) 
w.r.t. $\phi^\star$
after any $T\geq SA$ steps is upper bounded by
\beqan
\big(8\Diam^\star S^\star + 4\sqrt{S^\star}\big)\sqrt{A\log\big(\tfrac{48 S^\star A T^3}{\delta}\big)}\log\!\big(\tfrac{2T}{SA}\big)\\
\times \bigg(\sqrt{\big(AS + |\Phi|\big) T} + \big(AS + |\Phi|\big)\log\!\big(\tfrac{2T}{SA}\big)\bigg)\\
+ \big(\rho^\star + \Diam^\star \big)\big(AS+ |\Phi|\big)\log^2\!\big(\tfrac{2T}{SA}\big)\,
\eeqan
with probability higher than $1-\delta$, 
where $\rho^\star:=\rho^\star(\phi^\star)$, $S^\star:=S_{\phi^\star}$,
and $\Diam^\star:=\Diam(\phi^\star)$.

In particular, if for all $\phi\in\Phi$, $S_\phi\leq B$, then $S \leq B|\Phi|$ and hence with high probability 
\beqan
\Delta(\phi^\star,T) = \tilde O\big(\Diam^\star A B^{3/2}\sqrt{|\Phi| T} \big)\,.
\eeqan
\end{theorem}

\paradot{Comparison with the \texttt{BLB} algorithm}
Compared to the results obtained by \cite{MMR11} the regret bound in Theorem~\ref{thm:mainFinite}
has improved dependence of $T^{1/2}$ (instead of $T^{2/3}$) with respect to the horizon (up to logarithmic factors). Moreover, the
new bound avoids a possibly large constant for guessing the diameter of the MDP representation,
as unlike \texttt{BLB}, the current algorithm does not need to know  the diameter.
These improvements were possible since unlike \texttt{BLB} (which uses uniform exploration over all models,
and applies \texttt{UCRL2} as a ``black box") we employ optimistic exploration of the models, 
and do a more in-depth analysis of the ``\texttt{UCRL2} part'' of our algorithm.

On the other hand, we lose in lesser parameters: the multiplicative term in the new bound is
$S^\star A \sqrt{S} \leq S^\star A \sqrt{|\Phi|B}$
(assuming that all representations induce a model with no more than $S_\phi \leq B$ states),
whereas the corresponding factor in the bound of \cite{MMR11} is
$S^\star\sqrt{A|\Phi|}$. Thus, we currently lose a factor $\sqrt{AB}$.
Improving on the dependency on the state spaces is an interesting
question: one may note that the algorithm actually only chooses
models not much more complex (in terms of the diameter and the state space) 
than the best model. However, it is not easy to quantify this in terms of a concrete bound.

Another interesting question is how to reuse the information gained on one model for evaluation
of the others.	 Indeed, if we are able to propagate information to all models,
 a $\log(|\Phi|)$ dependency as opposed to the current $\sqrt{|\Phi|}$ seems
plausible. However, in the current formulation, a policy can be completely
uninformative for the evaluation of other policies in other models.
In general, this heavily depends on the internal structure of the models in $\Phi$. If all models induce state spaces
that have strictly no point in common, then it seems hard or
impossible to improve on $\sqrt{|\Phi|}$.

We also note that it is possible to replace the diameter in Theorem \ref{thm:mainFinite}
with the span of the optimal bias vector just as for the  \texttt{REGAL} algorithm \cite{BT09}
by suitably modifying the \AlgoName\ algorithm. However, unlike \texttt{UCRL2} and \texttt{OMS} 
for which computation of optimistic model and respective (near-)optimal policy can be performed by EVI,
this modified algorithm (as \texttt{REGAL}) relies on finding the solution to a constraint optimization problem,
 efficient computation of which is still an open problem.

%%%%%%%%
\section{Regret analysis of the \AlgoName~strategy}\label{sec:regret}
%%%%%%%%
The proof of Theorem \ref{thm:mainFinite} is divided into two parts. In Section~\ref{sub:test}, we first show that 
with high probability all Markovian state-representation models will collect sufficiently high reward according to the test in \eqref{eq:test}. 
This also means that the regret of any Markov model is not too large. This in turn is used in Section~\ref{sub:regret} to show that 
also the optimistic model employed by \AlgoName\ (which is not necessarily Markov) does not lose too much with respect to an optimal policy in an arbitrary Markov model. 
In our proof we use analysis similar to \cite{JOA10} and \cite{BT09}. 

%%%%
\subsection{Markov models pass the test in \eqref{eq:test}}\label{sub:test}
%%%%
Assume that $\phi_{k,j} \in \Phi$ is a Markov model.
We are going to show that $\phi_{k,j}$ will pass the
test on the collected rewards in \eqref{eq:test} of the algorithm at any step $t$ w.h.p.

{\bf Initial decomposition.}
First note that at time $t$ when the test is performed, we have $\sum_{s \in \cS_{{k,j}}}\sum_{a\in\cA} v_{k,j}(s,a) = \ell_{k,j} = t-t_{k,j}+1$, so that
 \beqa
 &&\hspace{-1cm}\ell_{k,j}\rho_{k,j} - \sum_{\tau=t_{k,j}}^t r_\tau  \nonumber \\
 &&\hspace{-0.5cm}= \sum_{s \in \cS_{{k,j}}}\sum_{a\in\cA} v_{k,j}(s,a)\Big(\rho_{k,j} - \hat{r}_{t_{k,j}:t}(s,a)\Big)\,,\label{eq:decomp}
 \eeqa
where $\hat{r}_{t_{k,j}:t}(s,a)$ is the empirical average reward collected for choosing $a$ in $s$
from time $t_{k,j}$ to the current time~$t$ in run $j$ of episode $k$.
Let $r_{k,j}^+(s,a)$ be the optimistic rewards of the model $M_{t_{k,j}}^+(\phi_{k,j})$ under policy~$\pi_{k,j}$
and $\v{P}^+_{k,j}$ the respective optimistic transition matrix.
Set $\v{v}_{k,j}:=(v_{k,j}(s,\pi_{k,j}(s)))_s\in \Real^{S_{{k,j}}}$, and let $\v{u}_{k,j}^{+}:=(u_{t_{k,j},\phi_{k,j}}^+(s))_s \in \Real^{S_{{k,j}}}$ be the state value vector given by EVI.
By \eqref{eqn:empevi}
and noting that $v_{k,j}(s,a)=0$ when $a\neq \pi_{k,j}(s)$ or $s\notin \mathcal S_{k,j}$, we get 
\beqa 
\lefteqn{\hspace{-0.55cm}\ell_{k,j}\rho_{k,j} - \!\!\!\! \sum_{\tau=t_{k,j}}^t \!\!\! r_\tau= \sum_{s,a}v_{k,j}(s,a)\big(\hat \rho^+_{k,j}(\phi_{k,j})-r_{k,j}^+(s,a)\big) } \nonumber\\
&&+\sum_{s,a}v_{k,j}(s,a)\big(r_{k,j}^+(s,a)-\hat{r}_{t_{k,j}:t}(s,a)\big)\nonumber\\
&\leq& \v{v}_{k,j}^\top\big(\v{P}_{k,j}^+- I\big) \v{u}_{k,j}^{+} \nonumber \\ 
&&+\sum_{s,a} v_{k,j}(s,a)\Big(r^+_{k,j}(s,a) - \hat{r}_{t_{k,j}:t}(s,a)\Big).\label{eqn:firstdecompo}
\eeqa
We continue bounding each of the two terms on the right hand side of \eqref{eqn:firstdecompo} separately.

{\bf Control of the second term.}
Writing $r(s,a)$ for the mean reward for choosing $a$ in $s$
(this is well-defined, since we assume the model is Markov), we have 
\begin{multline}
\hspace{-0.3cm}r^+_{k,j}(s,a) -  \hat{r}_{t_{k,j}:t}(s,a) = \big(r^+_{k,j}(s,a) - \hat r_{t_{k,j}}(s,a)\big)\\
 \quad + \big(\hat r_{t_{k,j}}(s,a) - r(s,a)\big) +  \big(r(s,a) -  \hat{r}_{t_{k,j}:t}(s,a)\big)\,.\nonumber
\end{multline}
The terms of this decomposition are controlled.
That is, using that $M(\phi_{k,j})$ is an admissible model according to \eqref{eqn:cond_r} with probability $1-\delta_{t_{k,j}}$
(by applying the results of measure concentration in Appendix C.1 of \cite{JOA10}
to the quantity $\hat r_{t_{k,j}}(s,a)$), and the mere definition of $r^+_{k,j}(s,a)$, and since $N_{t_k}(s,a) \leq N_{t}(s,a)$, 
we deduce that with probability higher than $1-\delta_{t_{k,j}}$,
\begin{eqnarray}
\lefteqn{\sum_{s,a} v_{k,j}(s,a)\Big(\big(r^+_{k,j}(s,a) - \hat r_{t_{k,j}}(s,a)\big)} \nonumber \\
     && + \big(\hat r_{t_{k,j}}(s,a) - r(s,a)\big) \Big) \nonumber\\
&\leq&\,\, 2\sum_{s,a} \frac{v_{k,j}(s,a)}{\sqrt{2 N_{t_k}(s,a)}}
 	\sqrt{\log\Big(\tfrac{2 S_{{k,j}} A t_{k,j}}{\delta_{t_{k,j}}}\Big)}\, \nonumber \\
&\leq& \,\, 	\sum_{s,a} \sqrt{2v_{k,j}(s,a)\log\Big(\tfrac{2 S_{{k,j}} A t_{k,j}}{\delta_{t_{k,j}}}\Big)}\,. \label{eq:2.1}
\end{eqnarray}
On the other hand, using again the results of measure concentration in Appendix C.1 of \cite{JOA10},
and that $v_{k,j}(s,a) \leq N_{t_k}(s,a) \leq t_{k,j}$,
we deduce by a union bound over $S_{{k,j}} A t_{k,j}$ events that
with probability higher than $1-\delta_{t_{k,j}}$ we get
\beqa
\lefteqn{\sum_{s,a} v_{k,j}(s,a)\Big(r(s,a) -  \hat{r}_{t_{k,j}:t}(s,a)\Big)} \nonumber\\
&&\leq\,\, \sum_{s,a} \frac{v_{k,j}(s,a)}{\sqrt{2 v_{k,j}(s,a)}}
\sqrt{\log\Big(\tfrac{2 S_{{k,j}} A t_{k,j}}{\delta_{t_{k,j}}}\Big)}\, \quad \nonumber\\
&&\leq\,\, \sum_{s,a} \sqrt{2v_{k,j}(s,a)
\log\Big(\tfrac{2 S_{{k,j}} A t_{k,j}}{\delta_{t_{k,j}}}\Big)}\,. \quad \label{eq:2.2}
\eeqa

{\bf Control of the first term.}  For the first term in \eqref{eqn:firstdecompo}, 
let us first notice that, since the rows of $\v{P}_{k,j}^+$ sum to $1$, 
$\big(\v{P}_{k,j}^+- I\big) \v{u}_{k,j}^{+}$ is invariant under a translation of the vector $\v{u}_{k,j}^{+}$.
In particular, we can replace $\v{u}_{k,j}^{+}$ with the quantity $\v{h}_{k,j}^+$,
where
\[ h_{k,j}^+(s) \eqdef u_{k,j}^+(s) - \min\big\{\,u_{k,j}^+(s)\,|\, s\in \cS_{{k,j}}\,\big\}\,. \]
Then, we make use of the decomposition
\begin{eqnarray}
\lefteqn{\v{v}_{k,j}^\top\big(\v{P}_{k,j}^+- I\big) \v{u}_{k,j}^{+} =}  \label{eq:first}\\
	&&\v{v}_{k,j}^\top\big(\v{P}_{k,j}^+- \v{P}_{k,j}\big) \v{h}_{k,j}^+
	  +\,\,\v{v}_{k,j}^\top\big(\v{P}_{k,j} - I\big) \v{h}_{k,j}^+\,,  \nonumber
\end{eqnarray}
where $\v{P}_{k,j}$ denotes the transition matrix corresponding to the MDP $M(\phi_{k,j})$
under policy $\pi_{k,j}$.
Since both matrices are close to the empirical transition matrix~$\hat { \v{P}}_{t_{k,j}}$
at time $t_{k,j}$, we can control the first term of this expression.

{\bf First part of the first term.}
Indeed, since $\sp_{k,j}^{+} = \|\v{h}_{k,j}^{+}\|_\infty$,
 we have for the first term in \eqref{eq:first},
using the decomposition $p_{{k,j}}^+(\cdot|s)- p_{k,j}(\cdot|s) =  \big(p_{{k,j}}^+(\cdot|s) - \hat p_{t_{k,j}}(\cdot|s)\big) + \big(\hat p_{t_{k,j}}(\cdot|s) -  p_{k,j}(\cdot|s)\big)$
together with a concentration result and the definition of $p_{{k,j}}^+$,
that  with probability higher than $1-\delta_{t_{k,j}}$
\begin{eqnarray}
\lefteqn{\v{v}_{k,j}^\top\big(\v{P}_{k,j}^+- \v{P}_{k,j}\big) \v{h}_{k,j}^{+}} \label{eq:firstfirst}\\
&=&  \sum_{s,a,s'} v_{k,j}(s,a) \Big(p_{{k,j}}^+(s'|s) - p_{k,j}(s'|s) \Big) h_{k,j}^+(s')  \nonumber\\
&\leq& \sum_{s,a} v_{k,j}(s,a) \,\big\| p_{{k,j}}^+(\cdot|s) - p_{k,j}(\cdot|s)\big\|_1 \cdot \big\|\v{h}_{k,j}^{+}\big\|_\infty \nonumber\\
&\leq& \sum_{s,a} 2\, v_{k,j}(s,a) \sqrt{\tfrac{2  \log(2^{S_{{k,j}}} S_{{k,j}} A t_{k,j}/\delta_{t_{k,j}})}{N_{t_k}(s,a)}}\, \big\|\v{h}_{k,j}^{+}\big\|_\infty \nonumber\\
&\leq& 2\,\sp_{k,j}^{+}\sum_{s,a}
	\sqrt{2 v_{k,j}(s,a) \log\Big(\tfrac{2^{S_{{k,j}}} S_{{k,j}} A t_{k,j}}{\delta_{t_{k,j}}}\Big)}\,.  \nonumber
\end{eqnarray}

{\bf Second part of the first term.}
The second term of \eqref{eq:first} can be rewritten using a martingale difference sequence.
That is, let $\mathbf{e}_s \in \Real^{S_{{k,j}}}$ be the unit vector with coordinates $0$ for all $s'\neq s$.
Following \cite{JOA10} we set $X_\tau \eqdef \big(p(\cdot| s_\tau,a_\tau) - \mathbf{e}^\top_{s_{\tau+1}}\big)\v{h}_{k,j}^{+}$ and get
\beqa
\lefteqn{\v{v}_{k,j}^\top\big(\v{P}_{k,j} - I\big) \v{h}_{k,j}^{+}}  \label{eq:e1}\\
&=& \!\!\!\! \sum_{\tau=t_{k,j}}^t \Big(p(\cdot| s_\tau,a_\tau) - \mathbf{e}^\top_{s_\tau}\Big)\, \v{h}_{k,j}^{+} \nonumber\\
&=& \!\!\!\!\Big(\mathbf{e}^\top_{s_{t+1}} - \mathbf{e}^\top_{s_{t_{k,j}}}
+ \!\!\sum_{\tau=t_{k,j}}^t \!\!\Big(p(\cdot| s_\tau,a_\tau) - \mathbf{e}^\top_{s_{\tau+1}}\Big) \Big) \, \v{h}_{k,j}^{+} \nonumber\\
&=& \!\!\!\!\sum_{\tau=t_{k,j}}^t X_\tau +  h_{k,j}^+(s_{t+1}) -  h_{k,j}^+(s_{t_{k,j}})\, \nonumber\\
&=& \!\!\!\!\sum_{\tau=t_{k,j}}^t X_\tau +  u_{k,j}^+(s_{t+1}) -  u_{k,j}^+(s_{t_{k,j}})\;.  \nonumber
\eeqa
Now the sequence $\{X_\tau\}_{t_{k,j}\leq \tau \leq t}$ is a martingale difference sequence with
\beqan
   |X_\tau| &\leq& \big\|p(\cdot| s_\tau,a_\tau) - \mathbf{e}^\top_{s_{\tau+1}}\big\|_1 \, \sp_{k,j}^{+} \,\leq\, 2\sp_{k,j}^{+}\,.
\eeqan
Thus, an application of Azuma-Hoeffding's inequality (cf.\ Lemma 10 and its application in \citet{JOA10}) 	
to \eqref{eq:e1} yields 
\beqa
&&\hspace{-0.5cm}\v{v}_{k,j}^\top\big(\v{P}_{k,j} - I\big) \v{h}_{k,j}^{+}  \nonumber\\
&&   \,\leq\,  2\sp_{k,j}^{+}\sqrt{2\ell_{k,j}\log(1/\delta_{t_{k,j}})}\,  +  \sp_{k,j}^{+} \, \label{eq:f1}
\eeqa
with probability higher than $1-\delta_{t_{k,j}}$.
Together with \eqref{eq:firstfirst} this concludes the control of the first term of~\eqref{eqn:firstdecompo}.

{\bf Putting all steps together.}
Combining \eqref{eqn:firstdecompo}, \eqref{eq:2.1}, \eqref{eq:2.2}, \eqref{eq:first}, \eqref{eq:firstfirst}, and \eqref{eq:f1},
we deduce that at each time~$t$ of run $j$ in episode $k$,
any Markovian model $\phi_{k,j}$  passes the test in \eqref{eq:test} with probability higher than $1-4\delta_{t_{k,j}}$.
Further, it passes all the tests in run~$j$ with probability higher than $1-4\delta_{t_{k,j}}2^j$.

%%%%
\subsection{Regret analysis}\label{sub:regret}
%%%%
Next, let us consider a model $\phi_{k,j} \in \Phi$, not necessarily  Markovian, that has been chosen at time $t_{k,j}$.
Let $t+1$ be the time when one of the three stopping conditions in the algorithm (lines 12, 17, and 19) is met.
Thus \AlgoName\ employs the model $\phi_{k,j}$ between $t_{k,j}$ and $t+1$, until a new model is chosen after the step $t+1$.
Noting that $r_\tau\in[0,1]$ and that the total length of the run is $(t+1) - t_{k,j} +1 = \ell_{k,j}+1$ 
we can bound the regret $\Delta_{k,j}$ of run $j$ in episode $k$ by
\beqan
\Delta_{k,j} &\eqdef& (\ell_{k,j}+1)\rho^\star - \sum_{\tau=t_{k,j}}^{t+1} r_\tau\\
&\leq&\ell_{k,j}\big(\rho^\star - \rho_{k,j}\big) + \rho^\star  +  \ell_{k,j}\rho_{k,j} - \sum_{\tau=t_{k,j}}^{t}r_\tau \,.
\eeqan
Since by assumption the test in \eqref{eq:test} has been passed for all steps $\tau \in [t_{k,j},t]$,
we have
\beqa\label{eq:epr}
   \Delta_{k,j} \leq \ell_{k,j}\big(\rho^\star - \rho_{k,j}\big) + \rho^\star  +  \lob_{k,j}(t),
\eeqa
and we continue bounding the terms of $\lob_{k,j}(t)$.

\paradot{Stopping criterion based on the visit counter}
Since $\sum_{s,a} v_{k,j}(s,a) = \ell_{k,j} \leq 2^j$, by Cauchy-Schwarz inequality
$\sum_{s,a} \sqrt{v_{k,j}(s,a)} \leq 2^{j/2}\sqrt{S_{{k,j}} A }$.
Plugging this into the definition~\eqref{eq:lob}
of $\lob_{k,j}$, we deduce from \eqref{eq:epr} that
\beqa
\lefteqn{\Delta_{k,j} \,\leq\, \ell_{k,j}\big(\rho^\star - \rho_{k,j}\big) + \rho^\star}  \label{eq:r1}\\
&&  + \sp_{k,j}^{+} + 2^{j/2} \sp_{k,j}^{+} c(\phi_{k,j};t_{k,j})  + 2^{j/2}c'(\phi_{k,j};t_{k,j}) \,. \nonumber 
\eeqa

\paradot{Selection procedure with penalization}
Now, by definition of the algorithm, for any optimal Markov model $\phi^\star$ defined in the statement of Theorem~\ref{thm:mainFinite},
whenever $M(\phi^\star)$  is admissible, i.e.\ $M(\phi^\star) \in \cM_{t_{k,j},\phi^\star}$ and was not eliminated during all
runs before run $j$ in episode $k$,
we have 
$\rho_{k,j} - \pen(\phi_{k,j};t_{k,j}) \geq \hat \rho_{k,j}^+(\phi^\star)- \pen(\phi^\star;t_{k,j}) \geq \rho^\star -  \pen(\phi^\star;t_{k,j}) - 2t_{k,j}^{-1/2}$,
or equivalently
\beqa
\lefteqn{\rho^\star - \rho_{k,j} \leq  \pen(\phi^\star;t_{k,j}) - \pen(\phi_{k,j};t_{k,j}) +2t_{k,j}^{-1/2}} \nonumber\\
&\leq&  2^{-j/2} c(\phi^\star;t_{k,j})\, \sp(\v{u}^+_{t_{k,j},\phi^\star})  \nonumber\\
&&      +  2^{-j/2} c'(\phi^\star;t_{k,j})  +  2^{-j} \sp(\v{u}^+_{t_{k,j},\phi^\star}) \nonumber \\
&&      -  2^{-j/2} c(\phi_{k,j};t_{k,j})\, \sp_{k,j}^{+} \nonumber\\
&&      - 2^{-j/2} c'(\phi_{k,j};t_{k,j}) - 2^{-j} \sp_{k,j}^{+} + 2t_{k,j}^{-1/2} . \quad \label{eq:long}
\eeqa
Noting that $\ell_{k,j}\leq 2^j$ and recalling that when $M(\phi^\star)$ is admissible,
the span of the corresponding optimistic model is less than the diameter of the true model, i.e.\ $\sp(\v{u}^+_{t_{k,j},\phi^\star})  \leq \Diam^\star$, see \cite{JOA10},
and we obtain from \eqref{eq:r1}, \eqref{eq:long}, and a union bound that
\beqa
\Delta_{k,j} &\leq& \rho^\star +  \Diam^\star +  2^{j/2} \Diam^\star c(\phi^\star;t_{k,j}) \nonumber\\
&& \,\,+\, 2^{j/2} c'(\phi^\star;t_{k,j}) + 2^{j+1} t_{k,j}^{-1/2} \,  \label{eq:dkj}
\eeqa
with probability higher than
\beqa
1-\sum_{k',j'; t_{k',j'}<t_{k,j}}4\delta_{t_{k',j'}}2^{j'} - 2\delta_{t_{k,j}}\,.\label{eq:proba}
\eeqa
The sum in \eqref{eq:proba} comes from the event that $\phi^\star$ passes all tests (and is admissible) for all runs in all episodes previous to time $t_{k,j}$,
and $2\delta_{t_{k,j}}$ comes from the event that $\phi^\star$  is admissible at time $t_{k,j}$.
We conclude in the following by summing $\Delta_{k,j}$ over all runs and episodes.

\paradot{Summing over runs and episodes}
Let $J_k$ be the total number of runs in episode $k$, and let $K_T$ be the total number of episodes up to time $T$.
Noting that $c(\phi^\star;t_{k,j})  \leq c(\phi^\star;T)$ and $c'(\phi^\star;t_{k,j})\leq c'(\phi^\star;T)$ as well as
using that $2 t_{k,j} \geq 2^j$ (so that $2^{j+1} t_{kj}^{-1/2} \leq 2\sqrt{2} \cdot 2^{j/2}$),
summing \eqref{eq:dkj} over all runs and episodes gives
\beqa
\lefteqn{\Delta(\phi^\star,T)\,=\, \sum_{k=1}^{K_T}\sum_{j=1}^{J_k} \Delta_{k,j} \leq  \big(\rho^\star + \Diam^\star \big)\sum_{k=1}^{K_T} J_k}  \label{eq:r2} \\
&&+ \Big(\Diam^\star c(\phi^\star;T) + c'(\phi^\star;T) + 2\sqrt{2}\Big) \sum_{k=1}^{K_T}\sum_{j=1}^{J_k} 2^{j/2}, \nonumber 
\eeqa
with probability higher than $1-\sum_{k=1}^{K_T}\sum_{j=1}^{J_k} 4\delta_{t_{k,j}} 2^j$, 
where we used a union bound over all events considered in \eqref{eq:proba} for the control of all the $\Delta_{k,j}$ terms,
avoiding redundant counts (such as the admissibility of $\phi^\star$ at time $t_{k,j}$).
Now, using the definition of $\delta_{t_{k,j}}$ and the fact that $2t_{k,j}\geq 2^j$, we get that 
\begin{multline*}
4\delta_{t_{k,j}}2^j = \frac{2^j\delta}{9t_{k,j}^2} \leq \frac{2^j\delta}{2t_{k,j}(t_{k,j} + 2^j)}\\
= \frac{\delta}{2t_{k,j}}  - \frac{\delta}{2(t_{k,j}+2^j)} \leq \sum_{t=t_{k,j}}^{t_{k,j}+2^j-1} \frac{\delta}{2t^2}\,,
\end{multline*}
where the last inequality follows by a series-integral comparison, using that $t\mapsto t^{-2}$ is a decreasing function. Thus, we deduce that the bound \eqref{eq:r2} is valid with probability at least $1- \sum_{t=1}^\infty \frac{\delta}{2t^2} \geq 1-\delta$
for all $T$,
and it remains to bound the double sum $\sum_{k}\sum_{j} 2^{j/2}$.

\paradot{From the number of runs..}
First note that by definition of the total number of episodes $K_T$ we must have
\begin{equation}\label{eq:sumrounds}
 T \geq    \sum_{k=1}^{K_T}\sum_{j=1}^{J_k-1}2^j
= \sum_{k=1}^{K_T}\Big(2^{J_k}-2\Big)\,,
\end{equation}
which implies also that we have the bound
\begin{equation*}
\sum_{k=1}^{K_T}\sum_{j=1}^{J_k} 2^j  =
2\sum_{k=1}^{K_T}\Big(2^{J_k}-2\Big) + 2K_T \leq 2T +2K_T.
\end{equation*}
Further, by Jensen's inequality we get
\beqa
 \sum_{k=1}^{K_T}\sum_{j=1}^{J_k-1} 2^{j/2} &\leq& \st \sqrt{ \sum_{k=1}^{K_T} J_k} \;\sqrt{\sum_{k=1}^{K_T}\sum_{j=1}^{J_k} 2^j  } \nonumber\\
 &\leq&  {\st \sqrt{ \sum_{k=1}^{K_T} J_k}} \;\sqrt{2T + 2K_T}\,.  \label{eq:er}
\eeqa

Now, to bound the total number of runs $\sum_{k=1}^{K_T} J_k$,
 using Jensen's inequality and \eqref{eq:sumrounds}, we deduce
\beqa
\hspace{-0.2cm}\sum_{k=1}^{K_T} J_k &=& \sum_{k=1}^{K_T} \log_2(2^{J_k}) \,\leq\,  K_T \log_2\Big( \frac{1}{K_T} \sum_{k=1}^{K_T}2^{J_k} \Big) \nonumber\\
\hspace{-0.2cm}&\leq& K_T \log_2\big( \tfrac{T}{K_T}+2 \big) \leq K_T \log_2\big( \tfrac{2T}{K_T} \big),\label{eq:rounds}
\eeqa
and thus it remains to deal with $K_T$.

\paradot{... to the number of episodes}
First recall that an episode is terminated when either the number of visits in some state-action pair $(s,a)$ has been doubled
(line 17 of the algorithm) or when the test on the accumulated rewards has failed (line 12). We know that with probability at
least $1-\delta$ the optimal Markov model is not eliminated from $\Phi$, while non-Markov models failing the test are deleted from $\Phi$.
Therefore, with probability $1-\delta$ the number of episodes terminated with a model  failing the test is upper bounded by $|\Phi|-1$.

Next, let us consider the number of episodes which are ended since the number of visits in some state-action pair $(s,a)$ has been doubled.
Let $K(s,a)$ be the number of episodes which ended after the number of visits in $(s,a)$
has been doubled, and let $T(s,a)$ be the number of steps in these episodes.
As it may happen that in an episode the number of visits is doubled in more than
one state-action pair, we assume that $K(s,a)$ and $T(s,a)$ count only the episodes/steps where $(s,a)$ is the
first state-action pair for which this happens.
It is easy to see that $K(s,a)\leq 1+\log_2 T(s,a) = \log_2 2T(s,a)$ for $T(s,a)>0$.
Then the bound $\sum_{s\in \cS}\sum_{a\in\cA} \log_2 2T(s,a)$ on the total number of these episodes is maximal under the constraint
$\sum_{s\in \cS}\sum_{a\in\cA} T(s,a)=T$
when $T(s,a)=\frac{T}{SA}$ for all $(s,a)$.
This shows that the total number of episodes $K_T$ is upper bounded by
\begin{equation}\label{eq:ep}
K_T \,\leq\,  S A \log_2\!\big(\tfrac{2T}{S A}\big) + |\Phi|-1
\end{equation}
with probability $1-\delta$, provided that $T\geq SA$.

\paradot{Putting all steps together}
Combining \eqref{eq:r2}, \eqref{eq:er} and \eqref{eq:rounds}
we get
$\Delta(\phi^\star,T) \leq \big(\rho^\star + \Diam^\star \big)K_T \log_2\!\big( \tfrac{2T}{K_T}\big)
+\big(\Diam^\star c(\phi^\star;T) + c'(\phi^\star;T) +2\sqrt{2}\big)\sqrt{2K_T \log_2\!\big( \tfrac{2T}{K_T}\big)\big(T+K_T\big)}$.
Hence, by \eqref{eq:ep} and the definition of $c$, $c'$,
the regret of \AlgoName~is, with probability higher than $1-\delta$, bounded by 
\beqan
\lefteqn{\Delta(\phi^\star,T) \,\leq\,  \big(\rho^\star + \Diam^\star \big)\big(S A + |\Phi|\big)\log^2_2(\tfrac{2T}{SA})} \\
&&\hspace{-0.6cm}+  \Big(\,\, 2\Diam^\star  \sqrt{2 S^\star A \log\!\big(\tfrac{2^{S^\star}24S^\star A T^3}{\delta}\big)} 
    + 2\Diam^\star \sqrt{2\log\!\big(\tfrac{24T^2}{\delta}\big)}\\
 &&\hspace{-0.6cm} +\, 2\sqrt{2 S^\star A \log\!\big(\tfrac{48 S^\star A T^3}{\delta}\big)} +  2\sqrt{2} \,\,\Big) \\
&&\hspace{-0.6cm}  \times \log_2\!\big(\tfrac{2T}{SA}\big)\Big(\!\sqrt{\big(S A + |\Phi|\big) 2T} \!+ \big(S A + |\Phi|\big)\log_2\!\big(\tfrac{2T}{SA}\big)\!\Big),
\eeqan
and we may conclude the proof with some minor simplifications.

%%%%%
\section{Outlook}\label{sec:disc}
%%%%%
The first natural question about the performance guarantees obtained is whether they are optimal. 
We know from the corresponding lower-bounds for learning  MDPs \cite{JOA10} that the dependence on $T$ we get for \AlgoName\  is indeed optimal.
Among other parameters, perhaps the most important one is the number of models $|\Phi|$; here we conjecture that the
$\sqrt{|\Phi|}$ dependence we obtain is optimal, but this remains to be proven.  Other parameters are the size of the action and state spaces for each model;
here we lose with respect to the precursor \BLB\ algorithm (see the remark after Theorem~\ref{thm:mainFinite}), and thus have room for improvement. It may be possible to obtain a better dependence for \AlgoName\ at the expense of
more sophisticated analysis.  Note, however, that so far there are no
known algorithms for learning even a single MDP that would have known optimal dependence on all these parameters.\\
Another important direction for future research is  infinite sets $\Phi$ of models; perhaps, countably infinite sets is the natural first step, with
separable --- in a suitable sense --- continuously-parametrized general classes of models being a foreseeable extension. A problem with the 
latter formulation is that one would need to formalize the notion of a model being close to a Markovian model and quantify the resulting
regret.

\section*{Acknowledgments}
This work was supported by the French National Research Agency (ANR-08-COSI-004 project EXPLO-RA),
by the European Community's Seventh Framework Programme (FP7/2007-2013) under grant agreement  270327 (CompLACS) and  216886 (PASCAL2),
 the Nord-Pas-de-Calais Regional Council and FEDER through CPER 2007-2013, the Austrian Science Fund (FWF): J~3259-N13, and
the Australian Research Council Discovery Project DP120100950, NICTA.

\end{document}